\documentclass{ifacconf}

\usepackage{graphicx}      
\usepackage{natbib}        
\usepackage{amssymb}
\usepackage{amsmath}
\usepackage{color}
\usepackage{booktabs}

\begin{document}
\begin{frontmatter}

\title{A Framework for Collision-Tolerant Optimal Trajectory Planning of Autonomous Vehicles} 


\author[First]{Mark L. Mote} 
\author[First]{Juan-Pablo Afman} 
\author[First]{Eric Feron}

\address[First]{Georigia Institute of Technology, Atlanta, GA 30332 USA        \\ (e-mail: \{mmote3, jafman3, feron\}@gatech.edu)}

\begin{abstract}                
Collision-tolerant trajectory planning is the consideration that collisions, if they are planned appropriately, enable more effective path planning for robots capable of handling them. A mixed integer programming (MIP) optimization formulation demonstrates the computational practicality of optimizing trajectories that comprise planned collisions. A damage quantification function is proposed, and the influence of damage functions constraints on the trajectory are studied in simulation. Using a simple example, an increase in performance is achieved under this schema as compared to collision-free optimal trajectories. 

\end{abstract}

\begin{keyword}
Trajectory planning, Autonomous vehicles, Optimal control, Predictive control, Impact
\end{keyword}

\end{frontmatter}

\section{Introduction}
Guidance, navigation and control is becoming more important than ever for most robotic applications. The tremendous increase in computational power of embedded systems has enabled the apparition of a wide range of ground and air vehicle platforms with tremendous capabilities and ever increasing autonomy. The path and trajectory planning problem is a well studied area. Reviews of the development of various techniques may be found in \cite{Tang,Kunchev,Raja}. The classical path planning paradigm involves finding a collision-free path between two given points, often while optimizing some cost function, usually based on time and/or energy. \cite{Raja} Similarly, the traditional trajectory planning problem seeks to find reference inputs for a collision free path between to points, while respecting the kino-dynamic properties of the vehicle, and most commonly minimizing time, energy, or jerk \cite{Gasparetto}. 


In contrast, physical contact is one of the primary modes of interaction among objects operating at human length scales. Contact can be used to transfer, redirect or dissipate energy and momentum, as well as accomplish specialized tasks requiring the manipulation of other bodies. In nature, one may observe the use of contact for more more effective navigation, such as swinging on branches, jumping to and from objects, or pushing off walls to redirect momentum. The authors believe that in certain cases, both the performance and capabilities of a vehicle can be improved by extending the set of allowable behaviors.
Hence, this paper focuses on replacing the binary constraint of collision freedom with a novel metric that embraces the possibility of collisions. 

While there has been ample work in robotics in terms of object manipulation and contact based navigation through walking or running robots, the literature on contact based navigation is scarce and virtually non-existent in the realm of optimal trajectory. There has been limited work in the area of low velocity contact interaction and navigation with UAVs \cite{Alexis,Marconi}, and contact based navigation in the form of bipedal robots \cite{Posa}. However, to our knowledge, no framework has been developed for performance optimization of a vehicles trajectory through collisions. 


The paper is organized in the following manner: First, we present the basic formulation and context of the problem, including the challenges related to developing an accurate collision model and damage quantification scheme. Secondly, the formulation is applied to a general 1-dimensional vehicle casted as a mixed integer programming problem. Finally, the performance of this approach
is evaluated and compared to a similar optimal planning strategy for a specific case of navigating to a single waypoint near a wall. The analysis is performed in Matlab \cite{Matlab} using the Gurobi\cite{Gurobi} solver for the MIP optimization. 




\section{Modeling Collisions}
The general two body collision problem can be stated as follows: given the geometric, inertial, material, and dynamic properties of two bodies before a collision, predict the dynamic properties of those bodies after the collision \cite{Cataldo}. High velocity contact is a complex phenomena which requires the consideration of many interacting variables, consequently these are very difficult to model with high accuracy. Fortunately there have been many methodologies developed to model collision behaviors in a wide range of scenarios \cite{Chatterjee}. 
Furthermore, we may obtain reasonable results at moderate complexity by leveraging several simplifying assumptions such as rigidity, perfect inelasticity, and other geometric simplifications such as flatness or symmetry.


\subsection{Constraining Damage} 

One particularly interesting feature of component design is the consideration of maximum loads that a component may experience during its operational envelope. We propose the use of a damage function to quantify and constrain undesirable collision events. Introducing damage as a constraining quantity in the optimization allows for a robust design that is capable of properly handling the maximum stresses and strains that a component will see under its typical operation.
As with the cost function, the damage function will be dependent on system properties and goals of the mission. However, it should typically come in the form of single collision metrics such as maximum force, impulse, or pressure, jerk, or from functions of an aggregate quantity such as energy dissipation over a time period \cite{mark}. Single collision metrics would most likely relate to the completion of a mission, while aggregate damage functions may be useful in extending a systems operational lifetime. With more complex systems, many parameters might emerge with relevance to the damage function, such as relative geometry, or the material at the point of contact.

\subsection{Preparing for Contact}
In order to make the best use of a collision tolerant system, we may engineer the vehicle itself to handle collisions in a favorable way. For example, elasticity may be introduced to a vehicles exterior in a time critical mission where momentum transfer is desired, in contrast one might introduce inelastic collision dampers in fuel optimizing vehicles where energy dissipation is favorable. Most likely vehicles will choose to benefit from some dedicated exterior module for collision handling. In this case it would be favorable to add constraints on the attitude at the time of collision, or add additional weighting to the damage function for off-angle collision scenarios.

\begin{figure}[tbp]
  \begin{center}
    \includegraphics[width=0.45 \textwidth]{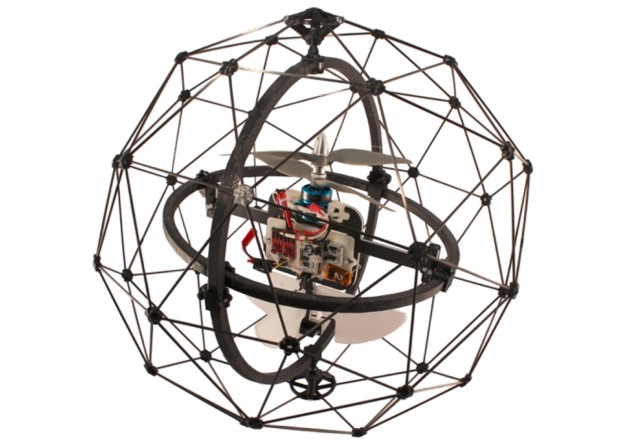}
    \caption{Gimball - a collision tolerant drone \cite{drone} }
    \label{fig:drone}
  \end{center}
\end{figure}

One of the greatest differences in collision tolerant trajectory planning as contrasted with previous methods is the profound importance that attitude orientation is expected to play; particularly when considering more sophisticated geometries. Inherently unstable vehicles such as multi-rotors may need to have very specific incidence angles at the time of collision to maintain stability. The accuracy of collision laws will become of more importance with these more complex systems. It may be of interest to create vehicles that have simplified geometries at the points of contact in order to improve the predictability and stability on impact, all while protecting the vehicle. One example of collision consideration in vehicle design can be seen in the Gimball UAV (Fig. \ref{fig:drone}), which features a spherical impact absorbing shell that rotates freely, allowing actuators and internal components to maintain orientation on contact. 

\section{Problem Formulation}
In this section we will develop the optimal trajectory problem formulation for a single vehicle in a 1-dimensional landscape modeled as a double integrator. We will make use of linear optimal programming to plan our trajectory, in particular we will build off of the MIP framework for trajectory generation in \cite{Schouwenaars1}. Aside from being highly efficient, the MIP framework provides a convenient platform for introducing and efficiently handling damage limitations as well as other dynamic and kinematic constraints in the optimization problem. Particularly, the inclusion of decision variables allows the encoding of logical decisions, as well as a way of modeling of the changing dynamics upon contact. We will use many of the techniques for hybrid modeling as MIPs that have been developed in \cite{Bemporad}.



\subsection{1-Dimensional Optimal Trajectory Generation}

In this section, the MIP problem will be formulated for the simplified case of a single vehicle with a 1-dimensional linear dynamic model. The vehicle will navigate from an initial location $x_{init}$ to a final goal location $x_g$. The trajectory will be calculated over the planning horizon $t=0,1,...,\tau$.
To ensure navigation between points, we will write constraints on the initial and final locations as,
\begin{equation}
x(0) = x_{init}
\end{equation}
\begin{equation}
x(\tau) = x_g
\end{equation}
The initial velocity $v(0)$ and initial acceleration $a(0)$ must also be defined for this problem. The transition between states is constrained to follow a set of linear dynamic equations. When the vehicle is not in a collision state, a double integrator model with acceleration input commands $u(t)=a(t)$ may be used. 
\begin{equation} \label{eq:cfdyn1}
 x(t+1) = x(t) + \Delta t v(t) + \frac{\Delta t^2}{2} a(t), \>\>\>\>\> t=0,...,\tau-1
\end{equation}

\begin{equation} \label{eq:cfdyn2}
v(t+1) = v(t) + \Delta t a(t), \>\>\>\>\> t=0,...,\tau-1
\end{equation}

The vehicle may be subject to input constraints of the form, 
\begin{equation} \label{eq:amax}
a(t) \leq a_{max}, \>\>\>\>\> t=1,...,\tau,
\end{equation}
\begin{equation} \label{eq:amin}
a(t) \geq a_{min}, \>\>\>\>\> t=1,...,\tau,
\end{equation}

Requirements for collision avoidance can be written as a constraint on the position of the vehicle $x(t)$. For example, given some obstacle placed at $x=x_w$, the following inequality defines feasible locations for the trajectory,
\begin{equation}
x(t) \geq x_w, \>\>\>\>\> t=1,...,\tau
\end{equation}
More sophisticated collision avoidance constraints for the higher dimensional cases can be found in \cite{Schouwenaars4}.

In order to modify this framework to consider collisions, a set of binary decision variables $\zeta(t)\in \{0,1\}$ must be introduced for each time-step $\Delta t$. Using a slack variable $M$, we can write constraints on $\zeta(t)$ such that it is forced to zero during non collision states ($x(t)\geq x_w$), and unity during collisions ($x(t)\leq x_w$).
\begin{equation} \label{eq:defwall1}
\>\>\>\>\>\>\>\>\>\>\>\>\>\>\>\>x(t) + M\zeta(t) \geq x_w, \>\>\>\>\> t=0,...,\tau
\end{equation}
\begin{equation}
-x(t) - M(1-\zeta(t)) \geq x_w, \>\>\>\>\> t=0,...,\tau
\end{equation}

Using this decision variable, we can switch between sets of constraints depending on whether the vehicle is in contact with the wall. In particular, we can use $\zeta(t)$ to deactivate the the input constraints (\ref{eq:amax}), (\ref{eq:amin}) on $a(t)$ and activate a collision law $a(t)=\mathcal{F}_{col}$ when the vehicle is in a contact state.
\begin{equation} \label{eq:colsat1}
a(t)-M\zeta(t)\leq a_{max}, \>\>\>\>\> t=0,...,\tau
\end{equation}
\begin{equation} \label{eq:colsat2}
a(t)+M\zeta(t) \geq a_{min}, \>\>\>\>\> t=0,...,\tau
\end{equation}
\begin{equation} \label{eq:collaw1}
a(t)-\mathcal{F}_{col}-M(1-\zeta(t))\leq 0 , \>\>\>\>\> t=0,...,\tau
\end{equation}
\begin{equation} \label{eq:collaw2}
a(t)-\mathcal{F}_{col}+M(1-\zeta(t))\geq 0 , \>\>\>\>\> t=0,...,\tau
\end{equation}
Note that the equations (\ref{eq:colsat1}), (\ref{eq:colsat2}), enforce the input saturation constraint when $\zeta(t)=0$ and equations  (\ref{eq:collaw1}), (\ref{eq:collaw2}) enforce the collision law when $\zeta(t)=1$. 

If it is desirable to limit the damage acting on the vehicle, then additional states $D(t)$ may be defined to represent this damage at each time step $t=0,1,...,\tau$. The decision variable can be used to assign values to the damage during contact iterations according to some damage function $D(t) = \mathcal{F}_{d}$.

\begin{equation} \label{eq:dl1}
D(t) - M\zeta(t) \leq 0, \>\>\>\>\> t=0,...,\tau
\end{equation}
\begin{equation} \label{eq:dl2}
D(t) + M\zeta(t) \geq  0, \>\>\>\>\> t=0,...,\tau
\end{equation}
\begin{equation} \label{eq:dl3}
D(t) - \mathcal{F}_{d} -  M(1-\zeta(t)) \leq 0, \>\>\>\>\> t=0,...,\tau
\end{equation}
\begin{equation} \label{eq:dl4}
D(t) - \mathcal{F}_{d} + M(1-\zeta(t)) > -\epsilon, \>\>\>\>\> t=0,...,\tau
\end{equation}

Likewise, a cumulative damage state $D_{total}$ can be defined as the sum of the individual damage values over the horizon,
\begin{equation} \label{eq:dl1}
D_{total} - D(0) - D(1) - ... - D(\tau) = 0
\end{equation}

Constraints may then be added on the damage, 
\begin{equation} \label{eq:dl5}
D(t) < D_{max}, \>\>\>\>\> t=0,...,\tau
\end{equation}
or the aggregate damage, 
\begin{equation} \label{eq:dltot}
D_{total} \leq D_{total,max}
\end{equation}

In addition to constraining the damage, it may also be desirable in some cases to penalize decisions to collide by incorporating the damage variables it into the cost function $J$.

\section{Implementation }
Employing the formulation developed in the previous section, we now develop a simulation experiment for comparing the \textit{collision-tolerant} optimization approach with a \textit{collision-free} optimal case. Similar to the collision-tolerant case, the linear program for generating the \textit{collision-free} optimal case trajectory with respect to a quadratic cost function $J$ is given by


\begin{equation}
\begin{array}{rlr} 
  \min_{a(t)}(J)&=\sum_{t=0}^{\tau}F_{0}\\
\\
s.t. 
 x(t+1) - \mathcal{G}_x(t) &=0, \>\>\> t=0,...,\tau &  (a)\\
 v(t+1)-\mathcal{G}_v(t) &=0, \>\>\> t=0,...,\tau  &(b)\\
  x(0)&=x_{init}     &(c)\\
  x(\tau)&=x_{g}   &(d)\\
  v(0)&=v_{init}   &  (e)\\
  v(\tau)&=v_{final}  &  (f)\\
  a(0)&=a_{init}    & (g)\\
  a(\tau)&=a_{final}  &  (h)\\
  a(t) &< a_{max}  &(i) \\
 a(t) &> a_{min}  &(j) \\
  x(t) &> 0,\>\>\> t=1,...,\tau & (k) \\
\end{array}\label{eq:Cost Minimizing Opt with Damage}
\end{equation}

where the terms $\mathcal{G}_x$ and $\mathcal{G}_v$ follow equations (\ref{eq:cfdyn1}), (\ref{eq:cfdyn2}) respectively. The mixed integer linear program for the collision-tolerant case is generated by replacing constraints (i)-(k) with the constraints (\ref{eq:defwall1})-(\ref{eq:dltot}). 

For the analysis, we begin by comparing the time-optimal solutions of each algorithm. The time-optimal trajectories for both optimization schemes are taken as the feasible trajectory with the lowest time horizon $\tau=\tau_min$. This value is found by setting the cost function to zero and iterating through $\tau$ until the minimum feasible solution is found. This search is done in MATLAB, which interfaces with Gurobi Optimization to solve for the trajectory.The set of initial and final conditions used in the simulation are provided in Table \ref{tab:IC}. 




\begin{table}[h!]
  \centering
  \caption{Vehicle's initial and final conditions}
  \label{tab:IC}
  \begin{tabular}{cc}
    \toprule
    Parameter & Value \\
    \midrule
    Initial Position & 10 m  \\
    Initial Velocity & 0 m/s\\
    Initial Acceleration & 0 m/s$^2$\\
    Final Position & 0.3 m\\
    Final Velocity & 0 m/s\\
    Final Acceleration & 0 m/s$^2$\\
    \bottomrule
  \end{tabular}
\end{table}

Similarly, Table \ref{tab:Constraint} provides the constraints imposed on the dynamics of the vehicle.
\begin{table}[h!]
  \centering
  \caption{Vehicle's Constraints}
  \label{tab:Constraint}
  \begin{tabular}{cc}
    \toprule
    Parameter & Value \\
    \midrule
    Max Acceleration & 6 m/s$^2$\\
    Max Velocity & 15 m/s\\
    \bottomrule
  \end{tabular}
\end{table}

\subsubsection{Results: Comparing trajectory planners}
\begin{figure}[h!]
\begin{center}
\includegraphics[width=9cm]{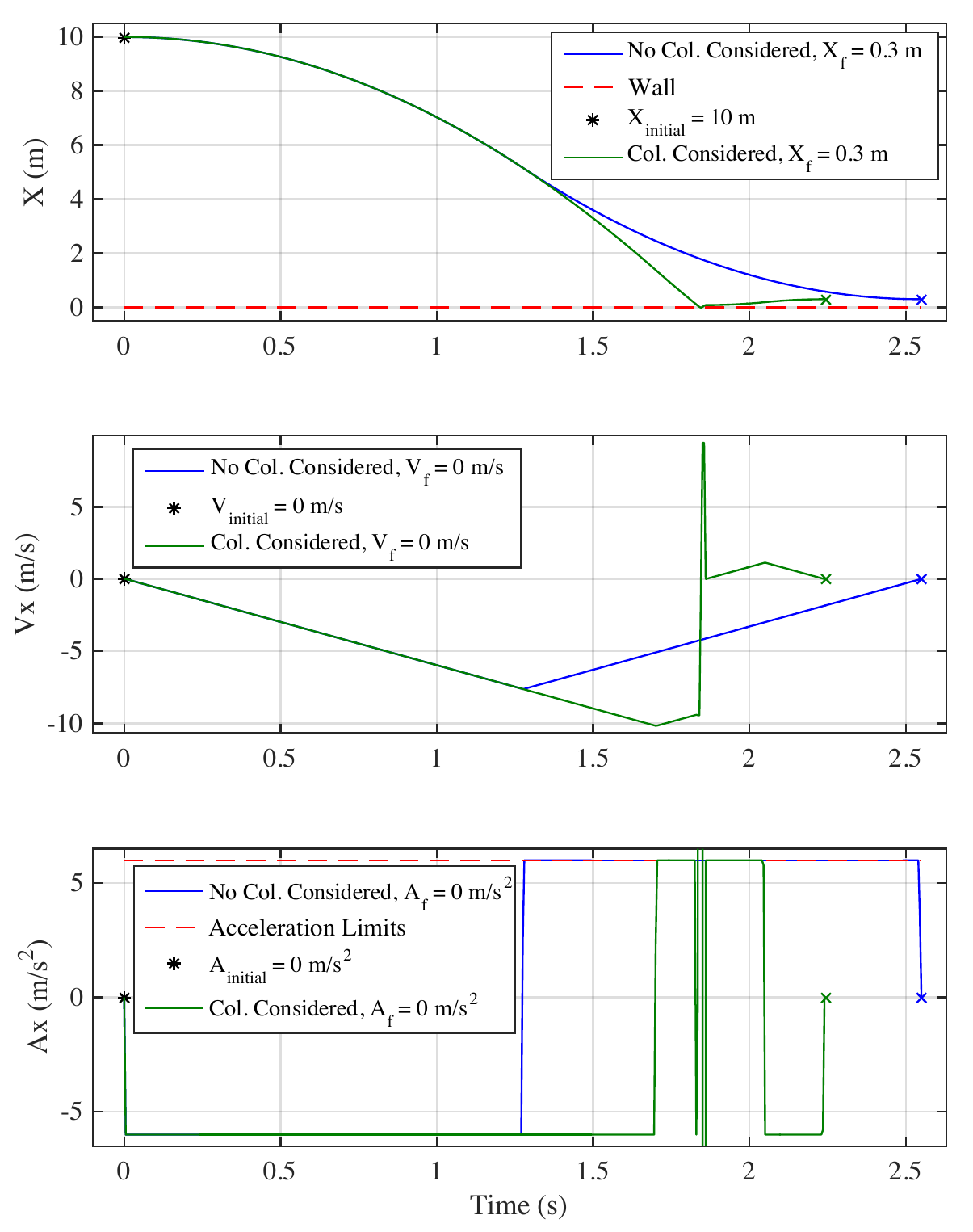}
\caption{Comparison of trajectory planners}
\label{comp}
\end{center}
\end{figure}

Figure \ref{comp} illustrates the comparison of an optimal trajectory of a vehicle, as it travels from its initial state to its final state, employing both collision-free and collision-tolerant optimization schemes. The possible interaction in this case is defined by a solid wall placed at $x=0$ meters. An inelastic collision was employed in this study, while the damage function was neglected.
From Figure \ref{comp} one can conclude that the results for the optimization which considers collisions outperforms the optimization which does not considers collisions. In fact, the collision-tolerant algorithm reaches its desired final state 13.59\% faster than its collision-intolerant counterpart. A breakdown of the results is provided in \ref{tab:table3}.

\begin{table}[h!]
  \centering
  \caption{Breakdown of Results}
  \label{tab:table3}
  \begin{tabular}{ccc}
    \toprule
    Test & Time (sec) & Difference (\%)\\
    \midrule
    No collision & 2.55 s & Baseline\\
    Collision &  2.245 s & 13.59\% faster\\
    \bottomrule
  \end{tabular}
\end{table}

\begin{figure}[h!]
\begin{center}
\includegraphics[width=8.5cm]{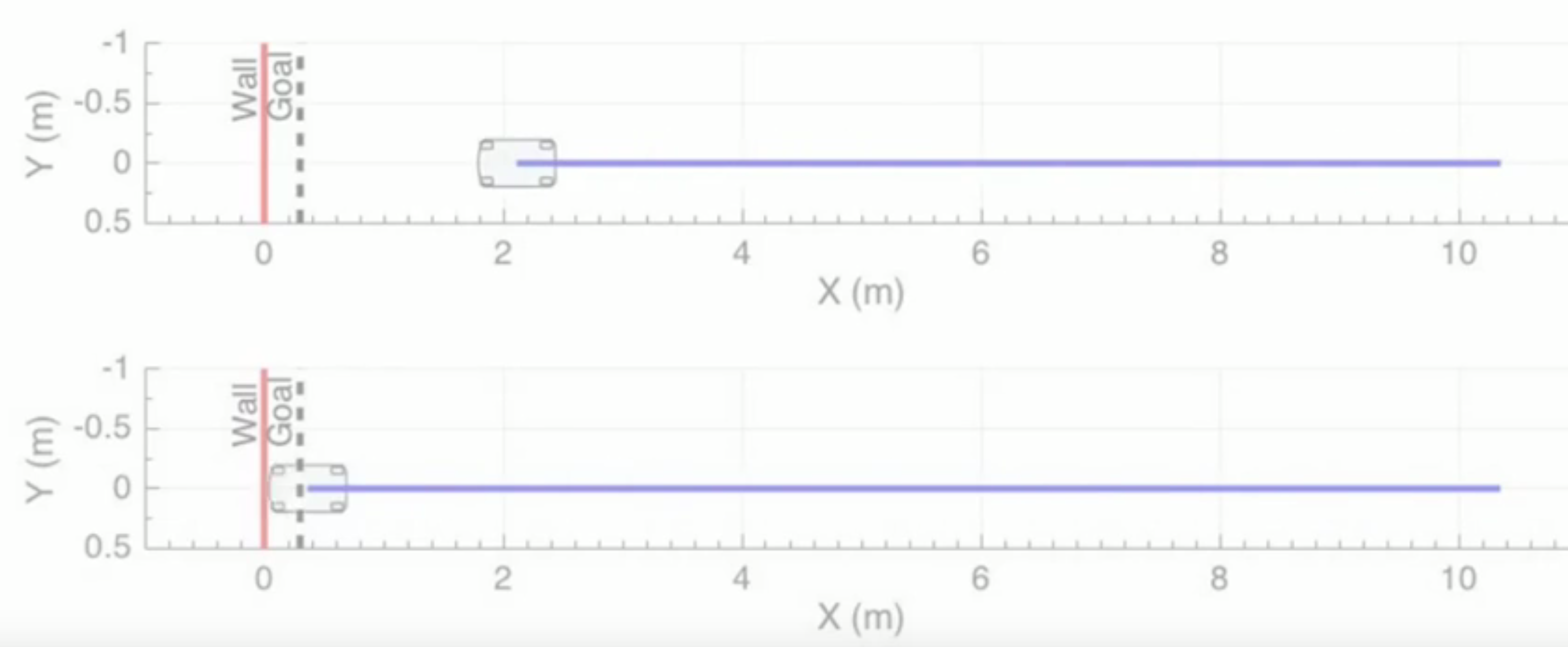}
\caption{https://youtu.be/rBrLK6xuXV8 Paste to browser to watch video}
\label{vid}
\end{center}
\end{figure}

A simulation employing the trajectories given by Figure \ref{comp} can be seen by following the link provided in Fig. \ref{vid}.

\newpage
\subsubsection{Results: Implementation of Damage}
Figure \ref{comp} illustrates the comparison of an optimal trajectory of a vehicle, as it travels from its initial state to its final state, employing both collision-free and collision-tolerant optimization schemes. The possible interaction in this case is defined by a solid wall placed at $x=0$ meters. An inelastic collision was employed in this study, and the damage function is now employed by introducing a constraint on the maximum velocity at impact, if a collision trajectory is sought.
\begin{figure}[h!]
\begin{center}
\includegraphics[width=8.5cm]{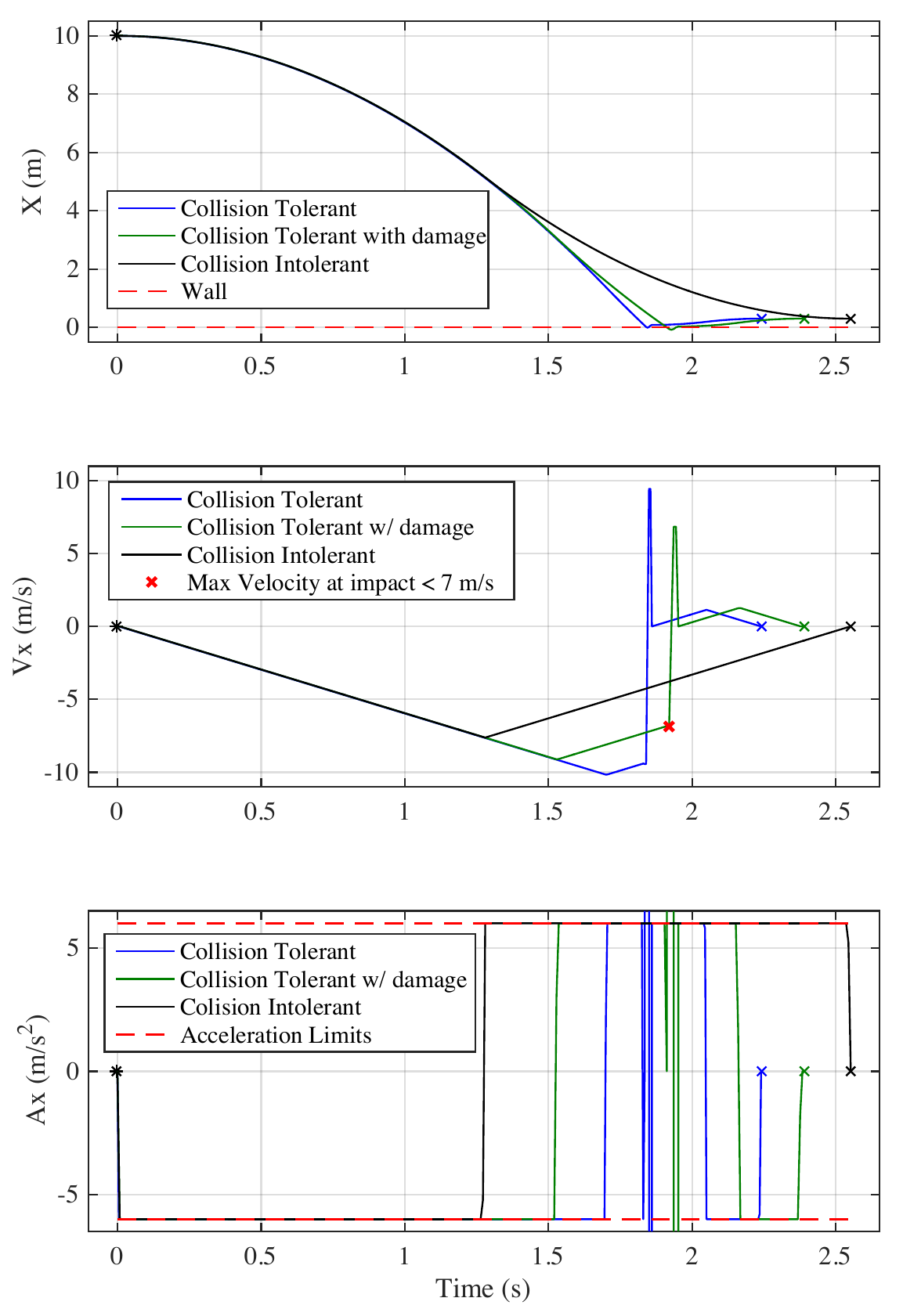}
\caption{Comparison of trajectory planners}
\label{dam}
\end{center}
\end{figure}
From Figure \ref{dam} one can conclude that the results for the optimization which considers collisions outperforms the optimization which does not considers collisions. A breakdown of the results is provided in \ref{tab:table4}.

\begin{table}[h!]
  \centering
  \caption{Breakdown of Results}
  \label{tab:table4}
  \begin{tabular}{ccc}
    \toprule
    Test & Time (sec) & Difference (\%)\\
    \midrule
   Collision Intolerant & 2.55 s & Baseline\\
    Collision Tolerant w/ Damage &  2.39 s & 6.27\% faster\\
    Collision Tolerant  & 2.25 s & 12.03\% faster\\

    \bottomrule
  \end{tabular}
\end{table}

\newpage
\subsubsection{Results: Comparison of Decision Making}
\begin{figure}[h]
\begin{center}
\includegraphics[width=9cm]{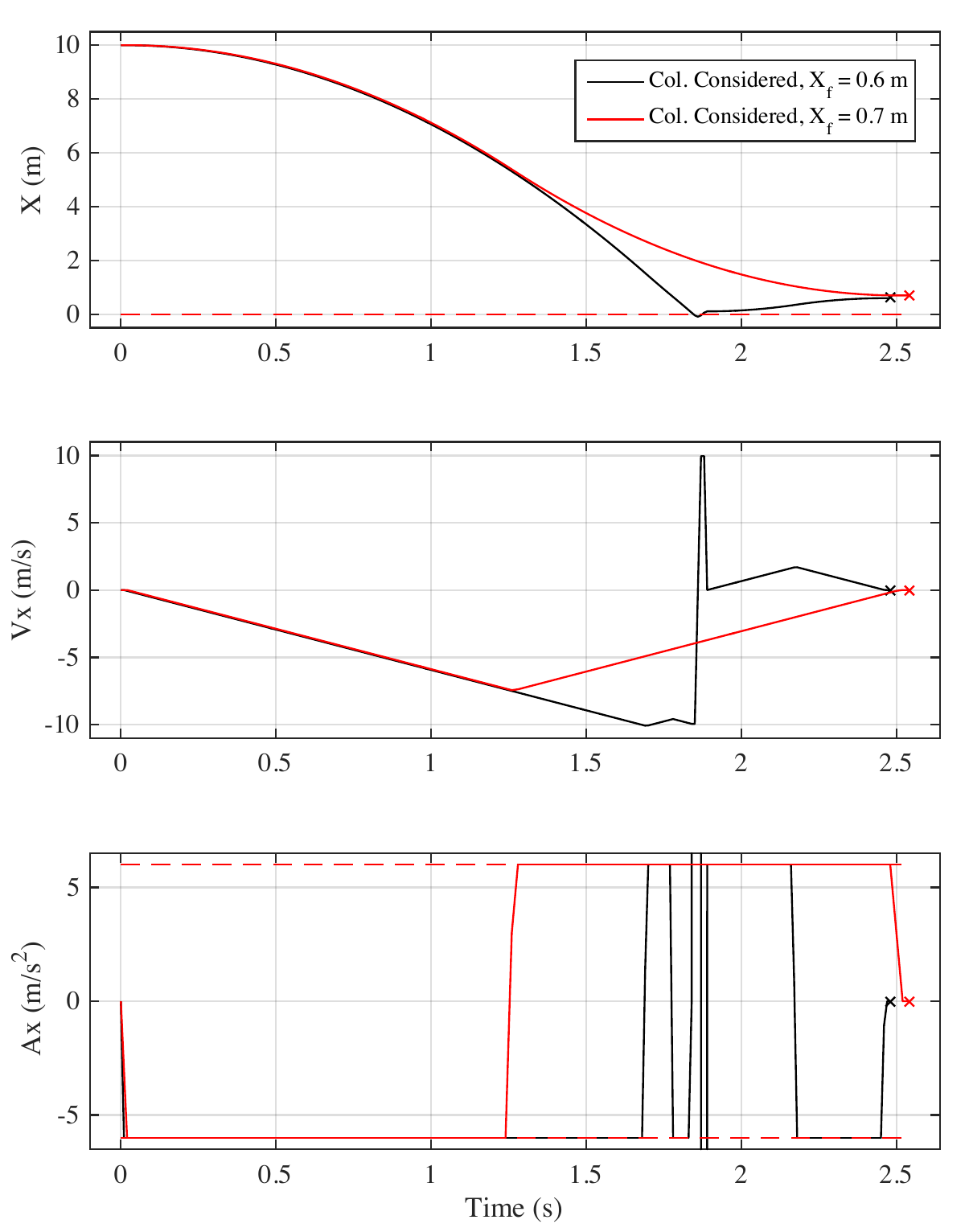}
\caption{Collisions considered but not necessarily pursued}
\label{des}
\end{center}
\end{figure}

The sensitivity of the collision-tolerant optimizer was also explored as a function of the final destination. Again, inelastic collisions were employed in this analysis. Figure \ref{des} illustrates this comparison, where the final position is changed by a small value $\Delta x=10$ cm.

It is observed that for a particular distance away from the wall, the collision-tolerant trajectory actually avoids the collision trajectory since it does not contribute to minimizing the cost.

\section{Future Work}
Now that the basic framework has been for formulated and validated in simulation for this simple scenario, we may begin to extend these strategies to higher dimensional cases, as well as more complex systems. Further research includes consideration of regulation to the optimal path, particularly at the wall. This work also shows potential in terms of beneficial contact based interaction in multi-agent systems. With the consideration of different geometries, it will be desirable to develop more detailed damage functions. Rather than considering damage solely as a constraint, it may be incorporated into the cost function to penalize inherently risky collision behavior. Indeed, one might extend this concept to formulate controllers for optimal damage mitigation in off nominal scenarios of vehicles not necessarily designed to crash. Specifically, this technique could be used to interrupt the primary controller of a vehicle if it is determined to be on an unwanted collision course. 

\section{Conclusion}
In this paper, Mixed Integer Programming is used to formulate a collision-tolerant optimal trajectory. This formulation is contrasted against its counterpart, where collisions are not tolerated, and it is shown that a collision tolerant trajectory planer can outperform the latter case. For a particular case studied in this work, a time savings of nearly $14\%$ was observed by employing of the collision-tolerant trajectory.   A damage function is included in the form of a maximum velocity constraint at impact. Although collision tolerant trajectories that take into account damage functions under-perform those that do not, they are shown to outperform optimal collision-free trajectories. Though collisions have been modeled as perfectly plastic interactions with the environment, extending the work to partially or fully elastic collisions, and adding various fixed or moving obstacles can be easily handled by the MIP framework.

\end{document}